%% file: paper.tex
\documentclass[a4paper,anonymous=false,USenglish]{lipics-v2021}

\title{Vehicle: Bridging the Embedding Gap in the Verification of Neuro-Symbolic Programs}
\titlerunning{Vehicle: Bridging the Embedding Gap} 

\author{Matthew L. Daggitt}{University of Western Australia, Perth, Australia}{}{}{}
\author{Wen Kokke}{Well-Typed, UK}{}{}{}
\author{Robert Atkey}{University of Strathclyde, Glasgow, UK}{}{}{}
\author{Ekaterina Komendantskaya}{Heriot-Watt and Southampton Universities, UK}{}{}{}
\author{Natalia Slusarz}{Heriot-Watt University, Edinburgh, UK}{}{}{}
\author{Luca Arnaboldi}{University of Birmingham, Birmingham, UK}{}{}{}

\authorrunning{M.L. Daggitt et al.}
\Copyright{Matthew L. Daggitt, Wen Kokke, Robert Atkey, Ekaterina Komendantskaya, Natalia Slusarz, Luca Arnaboldi}


\relatedversion{}
\supplement{}
\funding{The authors acknowledge support of the EPSRC grant AISEC: AI Secure and Explainable by Construction (EP/T026960/1, EP/T027037/1, EP/T026960/1). Slusarz acknowledges EPSRC DTA studentship awarded by Heriot-Watt University.  }
\acknowledgements{}

\nolinenumbers

\usepackage[T1]{fontenc}
\usepackage{graphicx}
\usepackage{amsmath}
\usepackage{amssymb}
\usepackage{subcaption}
\usepackage{syntax}
\usepackage[frozencache]{minted} 
\usepackage{multicol}

\usepackage{tikz}
\usetikzlibrary{math,arrows.meta}

\usepackage[textwidth=2.5cm,textsize=footnotesize,colorinlistoftodos]{todonotes}

\usepackage{agda}
\usepackage[utf8]{inputenc}
\input{agda-unicode.tex}

\input{macros.tex}

\EventEditors{Maribel Fern\'{a}ndez}
\EventNoEds{1}
\EventLongTitle{10th International Conference on Formal Structures for Computation and Deduction (FSCD 2025)}
\EventShortTitle{FSCD 2025}
\EventAcronym{FSCD}
\EventYear{2025}
\EventDate{July 14--20, 2025}
\EventLocation{Birmingham, UK}
\EventLogo{}
\SeriesVolume{337}
\ArticleNo{34}
\category{Invited Talk}
\supplementdetails[subcategory={Source Code}]{Vehicle Language, User Manual and Tutorials}{https://github.com/vehicle-lang/vehicle}
\relatedversiondetails[cite={DBLP:journals/corr/abs-2401-06379}]{Preliminary Draft Version}{https://arxiv.org/abs/2401.06379}

\begin{document}
	
\maketitle

\begin{abstract}
Neuro-symbolic programs, i.e. programs containing both machine learning components and traditional symbolic code, are becoming increasingly widespread. 
Finding a general methodology for verifying such programs is challenging due to both the number of different tools involved and the intricate interface between the ``neural'' and ``symbolic'' program components. 
In this paper we present a general decomposition of the neuro-symbolic verification problem into parts, and examine the problem of \emph{the embedding gap} that occurs when one tries to combine proofs about the neural and symbolic components. To address this problem we then introduce \vehicle -- standing as an abbreviation for a ``verification condition language'' -- an intermediate programming language interface between machine learning frameworks, automated theorem provers, and dependently-typed formalisations of neuro-symbolic programs.  
\vehicle{} allows users to specify the properties of the neural components of neuro-symbolic programs once, 
and then safely compile the specification to each interface using a tailored typing and compilation procedure.     
We give a high-level overview of \vehicle{}'s overall design, its interfaces and compilation \& type-checking procedures, and then demonstrate its utility by formally verifying the safety of a simple autonomous car controlled by a neural network, operating in a stochastic environment with imperfect information.
\end{abstract}

\section{Introduction}\label{sec:intro}

With the proliferation of neuro-symbolic systems that blend machine learning with symbolic reasoning, the formal verification of the reliability and safety of such systems is an increasingly important concern~\cite{lopez2023archcompainncs,seshia2022toward}.
Examples include: ensuring the correctness of decision-making software (e.g. insurance assessments~\cite{wuthrich2020bias}) where symbolic software delegate certain cases to trained neural models, and proving the safety 
of cyber-physical systems (e.g. cars~\cite{badue2021self} and drones~\cite{shi2019neural}) where neural agents must be proved safe with respect to a symbolic representation of the environment in which they act. 
Unfortunately, the non-interpretable nature of neural networks means that reasoning formally about these systems is significantly harder than reasoning about purely symbolic systems.
Despite this, the formal verification community has achieved notable successes, including the development of automatic theorem provers specialised for reasoning about neural network components in isolation~\cite{DBLP:journals/corr/abs-2412-19985}, and proving reachability results for neuro-symbolic systems~\cite{DBLP:conf/arch/LopezAFJL023}.  Nonetheless, current efforts verifying neuro-symbolic systems in general still face substantial challenges, with inconsistencies arising between different stages of training, implementation, verification and deployment~\cite{10.1007/978-3-031-91118-7}.

The contributions of this paper are as follows. In Section~\ref{sec:decomposition} we propose a general decomposition of the problem of training, constructing and verifying neural-symbolic systems. This decomposition reveals the difficulty of integrating proofs about the neural components with proofs about the symbolic components, which we call \emph{the embedding gap}. In particular, in the general case, we argue that neither interactive theorem provers nor existing automatic theorem provers (even those specialised in verifying neural networks) are suitable for carrying out this integration step. We illustrate the applicability of our analysis by describing a proof of temporal correctness of a simple autonomous car model operating in a non-deterministic, imperfect information environment.

In Section~\ref{sec:vehicle}, we present our \vehicle{} tool which is designed to enable the verification of neuro-symbolic systems by facilitating the decomposition we identified in the previous section and to close the embedding gap. The core of \vehicle{} is a high-level, dependently-typed language designed for writing specifications for the neural components of neuro-symbolic systems. It is optimised for expressivity and readability with support for tensors, neural networks, large datasets, first-class quantifiers and higher-order functions. The \vehicle{} compiler then translates these specifications to i) machine learning frameworks for training of the neural components, ii) automatic theorem provers for verification  of the neural components and iii) interactive theorem provers where proofs about the neural components can be integrated with proofs about the symbolic components. This paper also explains how, although Vehicle's dependent type-system is used directly when writing specifications in a limited fashion, its primary use is internally to translate code between the different backends and provide clear diagnostic error messages to users when their specifications cannot be compiled to a given backend.

\section{Analysing the Problem of Neural-Symbolic Verification}
\label{sec:decomposition}

\subsection{Decomposing the Problem}

We will begin by considering an abstract symbolic program $\system(\cdot)$, whose completion requires computing an unknown function~$\hypothesis : \problemSpace \rightarrow \resultSpace$, where $\hypothesis$ maps objects in the \emph{problem space}~$\problemSpace$ to those in the \emph{result space}~$\resultSpace$. The sets~$\problemSpace$ and~$\resultSpace$ may contain a mixture of discrete and continuous components, and crucially are semantically rich, by which we mean they refer to quantities interpretable by humans (e.g. measurements in real world units, images, text etc.).

As $\hypothesis$ is a complex function, the goal is to train a neural network to approximate it. However, neural networks can only learn functions between continuous spaces and perform best when the input data is normalised, whereas in general $\problemSpace$ and $\resultSpace$ may contain a mix of discrete and unnormalised continuous values. The standard approach is to construct an embedding function $e : \problemSpace \rightarrow \real^m$ and an unembedding function $u : \real^n \rightarrow \resultSpace$ that map the semantically meaningful objects to and from values in a continuous vector space, and then train a neural network~${f : \real^m \rightarrow \real^n}$ such that $u \circ f \circ e \approx \hypothesis$. We will refer to $\real^m$ and $\real^n$ as the \emph{input space} and \emph{output space} respectively. Unlike objects in the problem/result spaces, in general, embedded objects in the input/output spaces are unitless, normalised quantities and are therefore not semantically interpretable. Furthermore, for real systems $m$ and $n$ may be very large, e.g. for image classification networks $m$ will correspond to the number of pixels. 

The completed neuro-symbolic program is then modelled as $\system(u \circ f \circ e)$. Our end goal is to prove that~$\system(u \circ f \circ e)$ satisfies some property~$\systemProperty$, which we will refer to as the \emph{system property}. The natural way to proceed is to establish a \emph{solution property}~$\solutionProperty$ and a \emph{network property}~$\networkProperty$ such that the proof of~$\systemProperty$ is decomposable into the following three lemmas:
\begin{align}
    \label{eq:systemProperty}
    \forall h.~ \solutionProperty(h) \Rightarrow
    \systemProperty(\system(h))
    \qquad \qquad \qquad \qquad \qquad \qquad \qquad \qquad \qquad \qquad \\
    \label{eq:solutionProperty}
    \forall g.~ \networkProperty(g) \Rightarrow \solutionProperty(u \circ g \circ e)
    \qquad \qquad \qquad \qquad \quad  \\ 
    \label{eq:networkProperty}
    \networkProperty(f)
\end{align}
i.e. Lemma~\ref{eq:systemProperty} proves that the system satisifes the property~$\systemProperty$ for \emph{any} implementation of~$\hypothesis$ that satisfies~$\solutionProperty$. Crucially, this property requires only reasoning about the symbolic portion of the system. Lemma~\ref{eq:solutionProperty} links the symbolic and neural components of the proof by proving that if any network satisfies~$\networkProperty$ then when composed with the embedding functions it satisfies~$\solutionProperty$. Finally Lemma~\ref{eq:networkProperty} proves that the actual concrete network~$f$ obeys the network property~$\networkProperty$. Together they can be composed in the obvious way to show that the neuro-symbolic program~$\system(u \circ f \circ e)$ obeys the program property~$\systemProperty$.

\subsection{The Embedding Gap}\label{subsec:egap}

We now discuss how we can implement this proof strategy, starting with finding a suitable property~$\solutionProperty$ and proving Lemma~\ref{eq:systemProperty} which reasons about the symbolic component of the system. Determining what property~$\solutionProperty$ should be will usually require deep expertise in the problem domain, but fundamentally requires no new insights or methodology as we can rely on insights from the formal verification community which has many decades of experience in decomposing proofs about symbolic systems down into constituent parts. Likewise, once $\solutionProperty$ has been found, the community is well placed to prove results of this form using a variety of powerful interactive theorem provers (ITPs)~(e.g.~ \cite{barras1997coq,norell2008dependently}). 

We will come back to methods for finding a suitable property~$\networkProperty$ and proving Lemma~\ref{eq:solutionProperty} later. Instead we turn our attention to Lemma~\ref{eq:networkProperty}, the proof about the neural component of the system. Assuming one does have a suitable~$\networkProperty$, experience has shown that proving properties about neural networks directly in an ITP is challenging.
 The first issue is that the modular reasoning that ITPs excel at is not well suited to the non-interpretable and semantically non-compositional nature of neural networks.
Furthermore the sheer size of the networks, often millions or billions of parameters~\cite{Hughes23,yuan2021florence}, make even representing the network, let alone proving anything about it, impractical in an ITP.
For example, the largest neural network verified to date purely in an ITP is a few hundreds of  neurons (or a few thousand of weights)~\cite{bagnall2019certifying,BruckerS23,de2022use,DesmartinPKD22}.
In contrast, the automated theorem prover (ATP) community has been significantly more successful at proving properties of the form of Lemma~\ref{eq:networkProperty}. Starting with Reluplex~\cite{katz2017reluplex}, the community has rapidly developed highly specialised SMT and abstract interpretation-based solvers which are capable of verifying properties of neural networks with up to millions of neurons~\cite{katz2019marabou,muller2022prima,wang2021beta,zhang2022general}. There is, however, a further consideration. Unlike conventional software which is usually at least morally (if not actually) correct at verification time, neural networks often struggle to learn property $\networkProperty$ from data alone~\cite{szegedy2013intriguing}.
Consequently, we also need property $\networkProperty$ to influence the training of the network $f$, e.g. using techniques such as differentiable logic~\cite{FischerBDGZV19} and linear temporal logic-based reward functions~\cite{hasanbeig2023certified}.

This leaves the problem of how to link the proofs about the symbolic and the neural components of the system, by finding a suitable property~$\solutionProperty$ and proving Lemma~\ref{eq:solutionProperty}. 
Firstly, even finding a suitable~$\networkProperty$ is difficult as it refers to a semantically uninterpretable input and output spaces. 
This strongly suggests that we need a method of automatically deriving it from $\solutionProperty$ and the embedding functions.
Suppose we did have such an automatic procedure. ATPs are ill-suited to proving Lemma~\ref{eq:solutionProperty}, as they are not designed to reason about a) the discrete components present in $\problemSpace$ and $\resultSpace$ and b) the arbitrary computation present in the embedding functions~$u$ and~$e$. Unfortunately, ITPs are equally ill-suited to proving Lemma~\ref{eq:solutionProperty} it would require the user to manually write down and then reason about property~$\solutionProperty$ in the ITP directly (remember~$\solutionProperty$ is often uninterpretable
and scales with the size of the embedding space i.e. potentially tens of thousands of inputs). We call this lack of practical methodology for establishing the results that link the symbolic and neural components of the proof the \emph{embedding gap}.

Given the analysis above, it is clear that in general to construct and verify a neuro-symbolic program, we need machine learning frameworks, ATPs and ITPs to work together. 
Unfortunately, usually each of these have their own specialised input formats and semantics and currently the default approach is to write out the specification three times for the three different tools. This is deeply suboptimal as it requires an informal judgement that each of the three representations encode the same property.

\subsection{Our Vision of the Solution}
\label{sec:vision}

\begin{figure}[t]
	\centering
    \input{figures/specification-diagram.tex}
    \caption{The architecture of \vehicle{} for neuro-symbolic program verification. Dashed lines indicate information flow and solid lines automatic compilation.}
    \label{fig:specification-diagram}
\end{figure}

Figure~\ref{fig:specification-diagram} shows our vision of a tool that overcomes these problems and enables the \textit{general} verification of neuro-symbolic systems. 
Firstly, users should verify Lemma~\ref{eq:systemProperty} about the symbolic component of the system using whichever existing ITP best meets their needs. 
Next the user expresses the specification~$\solutionProperty$ and the embedding functions~$e$~and~$u$ in terms of the semantically-meaningful problem space using a suitable domain-specific language (DSL).
This specification of $\solutionProperty$ is then automatically compiled down to representations of $\networkProperty$ in the semantically uninterpretable input/output spaces suitable for i) training the network in the user's machine learning framework of the choice and ii) verifying the output of training using an appropriate ATP. Once a network $f$ has been trained that the ATPs can prove satisfies $\networkProperty(f)$, the proof of $\solutionProperty(u \circ f \circ e)$ should be constructed automatically and returned to the ITP.

There are several advantages to such a hypothetical tool. Firstly, the user only has to express the specification once and the tool automatically generates provably equivalent representations suitable for each of the backends.
Secondly, the specifications of the neural component $\solutionProperty$ are written in terms of the semantically meaningful problem and result spaces $\problemSpace$ and $\resultSpace$. This means they can be read and checked by experts in the problem domain who may not know anything about machine learning.

\section{A Concrete Example}
\label{sec:example}

In Section~\ref{sec:vehicle} we will introduce our tool \vehicle{} which implements a large proportion of our vision from Section~\ref{sec:vision}. However, before we do so, we will now give an example of a simple, concrete verification problem that illustrates our proposed decomposition described in Section~\ref{sec:decomposition} and will assist our explanations of \vehicle{}'s operation in Section~\ref{sec:vehicle}.

\begin{figure}[t]
    \centering
    \includegraphics[width=.5\textwidth]{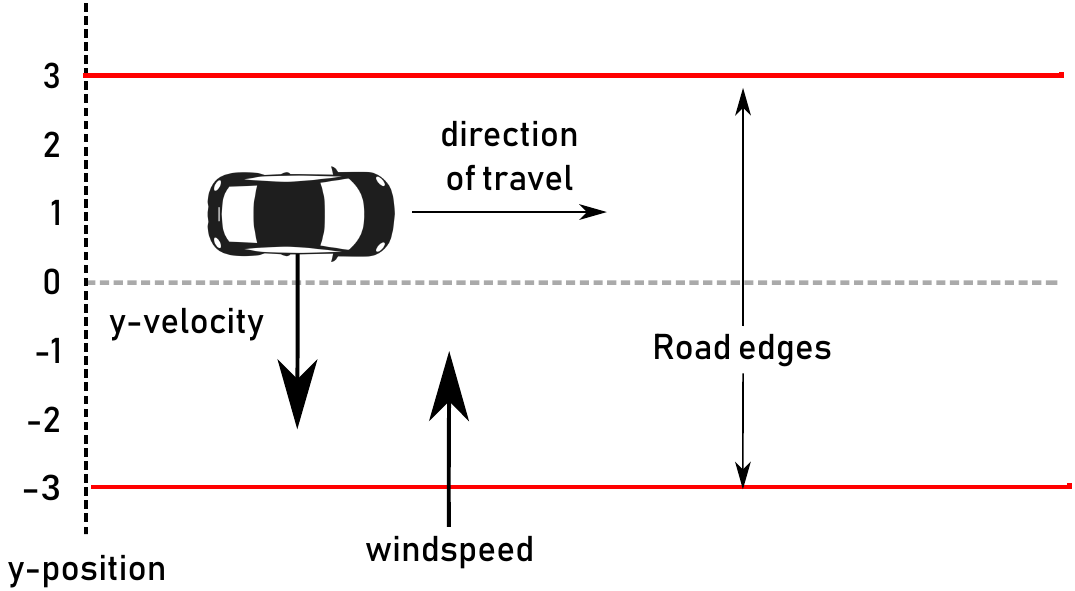}
    \caption{A simple model of an autonomous car compensating for a cross-wind.}
    \label{fig:wind-controller}
\end{figure}

As illustrated in Figure~\ref{fig:wind-controller}, we use a modified version of the verification problem presented by Boyer, Green and Moore~\cite{boyer1990use}. An autonomous car is travelling along a straight road of width 6 parallel to the x-axis, with a varying cross-wind that blows perpendicular to the x-axis. The car has an imperfect sensor that provides noisy measurements of its position on the y-axis, and can change its velocity with respect to the y-axis in response. The car's controller takes in both the current sensor reading and the previous sensor reading and its goal is to keep the car on the road.
The desired system safety property that we would like to prove is as follows:
\begin{center}
\emph{If the wind-speed never shifts by more than 1 per unit time and the sensor is never off by more than 0.25 units then the car will never leave the road.}
\end{center}
Note that this control problem involves both stochasticity via the fluctuating wind-speed and imperfect information via the error on the sensor reading.

\subsection{Symbolic Component} 

In order to prove the system property above, it is necessary to first construct a symbolic model of the behaviour of the system (i.e., $s(\cdot)$  in our analysis in Section~\ref{sec:decomposition}).
We discretise the model as in \cite{boyer1990use}, and then formalise it in Agda, an interactive theorem prover. As discussed in Section~\ref{sec:vision}, neither the discretisation nor the use of Agda are relevant to the central proposal of this paper. We could equally have chosen to create a continuous model of the system based on differential equations in alternative systems such as Rocq or KeYmaera~X.
 
The state of the system consists of the current wind speed, the position and velocity of the car and the most recent sensor reading. 
An oracle provides updates in the form of observations consisting of the shift in wind speed and the error on the sensor reading. 
The third component is a controller that takes as input the current and previous sensor readings and produces a recommended change in velocity:

\begin{minipage}{\linewidth}
\begin{multicols}{2}
    \input{agda/state.tex}
    \vfill\null
    \columnbreak
    \input{agda/observation.tex}
    \vspace{-3em}
    \input{agda/controller-type.tex}
\end{multicols}
\end{minipage}

\noindent Using these components, we can define the evolution of the system as:
\input{agda/nextState.tex}
\vspace{-2em}
\input{agda/finalState.tex}

\noindent Given suitable encodings of \AgdaFunction{ValidObservation} and \AgdaFunction{OnRoad}, the system safety property (i.e., $\systemProperty$ in our analysis in Section~\ref{sec:decomposition}) can be formalised as follows:
\input{agda/finalState-onRoad.tex}

\noindent This statement can be proved in Agda by induction over the list of observations, and can be found in the supplementary material. The proof crucially requires the \AgdaFunction{controller} to satisfy the following property ($\solutionProperty$ in our analysis in Section~\ref{sec:decomposition}):
\input{agda/controller-lemma.tex}

\noindent This says that if both the current and previous sensor readings say that the car is less 3.25 metres from the centre of the road, then the sum of the output of the controller and twice the current sensor reading minus the previous sensor reading must be less than 1.25.
The goal is to implement the function \AgdaFunction{controller} with a neural network that provably satisfies \AgdaFunction{controller-lemma}.

\subsection{Neural Component}

As this is a simple control problem, we choose 3 densely connected layers as the architecture for our neural network controller.
In terms of the generic problem decomposition we described in Section~\ref{sec:decomposition}, the embedding function $e$ normalises the semantically meaningful problem space inputs measured in metres in the range $[-4, 4]$ to the range $[0,1]$ in the embedding space.
The unembedding function $u$ is the identity function. 
Again, as discussed in Section 2.3, neither the choice of architecture nor the choice of embedding functions are relevant to the central proposal in the paper.

\section{The \vehicle{} Tool}
\label{sec:vehicle}

\begin{figure}[t]
    \input{figures/vehicle-controller-spec.tex}
    \caption{The safety property for the car's neural network controller expressed in \vehicle{} surface syntax .}
    \label{fig:vehicle-controller-spec}
\end{figure}

Our \vehicle{} tool implements the vision we described in Section~\ref{sec:vision}: users are provided a simple domain-specific language
with types to write a single specification about the neural component of the neuro-symbolic system (see Figure~\ref{fig:vehicle-controller-spec}). \vehicle{} then compiles this specification into forms suitable for both training and verification and can export the verified specification to ITPs. \vehicle{} can be installed by running ``\texttt{pip install vehiclepy}'',
which provides both a Python and a command line interface. A user manual and tutorials can be found online~\cite{FoMLAS2023,vehicle2024}.
 
\subsection{Specification Language}
\label{sec:vehicle-language}

\begin{figure}[t]
    \input{figures/vehicle-syntax.tex}
    \caption{Syntax for the core calculus of \vehicle{}}
    \label{fig:vehicle-syntax}
\end{figure}

The Vehicle Condition Language (\VSL{}) is designed for writing high-level, problem space specifications for neural networks.
At its core is a dependently-typed {$\lambda$-calculus} extended with operations for logic, arithmetic and manipulating tensors.
The abstraction capabilities of the $\lambda$-calculus enable users to write modular and reusable specifications, while the dependent types allow tensor dimensions to be tracked at the type level, preventing common specification errors such as dimensions mismatches and out-of-bounds indexing. A standard instance resolution mechanism allows for the overloading of operators in the surface syntax.
The syntax of the core language is shown in Figure~\ref{fig:vehicle-syntax}.

A \vehicle{} program consists of a list of declarations.
There are four non-standard declaration types in \VSL{} that describe how a specification links to the outside world.
Firstly, \texttt{@network} declarations introduce an abstract neural network into scope. Next \texttt{@parameter} and \texttt{@dataset}  declarations can be used to introduce external values that either may not be known in advance or are too big to represent directly in the specification.
Crucially, only the user only needs to provide the types of these declarations when writing the specification.
As discussed in Section~\ref{sec:implementation}, the user provdies the compiler with their actual implementation or value at compile time.
Finally, \texttt{@property} declarations are used to explicitly designate a constraint the neural networks are expected to satisfy. A complete description of the syntax is available in the user manual~\cite{vehicle2024}.

Figure~\ref{fig:vehicle-controller-spec} illustrates how the specification, $\solutionProperty$, for the car controller from Section~\ref{sec:example} is expressed in \vehicle{}.
The embedding function $e$ is represented by the ``normalise'' function on Lines~11-12.
Crucially, the safety conditions specified on Lines~14-20 are written in terms of the problem space, using units such as metres, rather than being expressed in the embedding space. This makes them interpretable and meaningful to readers of the specification.

\subsection{The Vehicle Compiler}
\label{sec:implementation}

The \vehicle{} compiler translates \VSL{} specifications into formats suitable for three different backends: Tensorflow loss functions  for training, Marabou queries for verification and Agda code for integration with the proof about the surrounding symbolic system. The compiler is implemented in Haskell, and its overall architecture is illustrated in Figure~\ref{fig:vehicle-type-checker}.

\begin{figure}[t]
\begin{tikzpicture}
\tikzstyle{typing}=[draw, rectangle, align=center, fill=blue!20];
\tikzstyle{error}=[draw, rectangle, align=center, fill=red!20, text width=1.5cm];
\tikzstyle{output}=[draw, rectangle, align=center, fill=green!20, text width=1.8cm];

\tikzmath{
	\agdaY=2.5;
	\linErrorY=1.25;
	\atpY=0;
	\polErrorY=-1.25;
	\lossY=-2.5;
	\typeX=2;
	\backendX=11;
	\outputX=8;
	\errorX=4.5;
	\errorCheckX=7.3;
	\errorMessageX=10.0;
}

\node (start) at (-2,\atpY) {};

\node[rectangle,draw] (scope) at (-0.75,\atpY) {\textbf{Scoping}};

\node[typing, text width=2.5cm] (type) at (\typeX,\atpY) {\textbf{Standard type-checking}};


\coordinate (normalise) at (5,\atpY) {};

\node[rectangle,draw] (queries) at (\outputX,\atpY) {\textbf{Queries}};

\node (endqueries) [output] at (\backendX,\atpY) {\textbf{Marabou}};

\draw [->] (start) edge (scope);
\draw [->] (scope) edge (type);
\draw [->] (type) edge node [above, pos=0.7]{verification} (queries);
\draw [->] (queries) edge (endqueries);


\node[rectangle,draw, align=center, text width=1.8cm] (linerror) at (\errorX,\linErrorY) {\textbf{Linearity error}};

\node[typing, text width=2.5cm] (lintype) at (\errorCheckX,\linErrorY) {\textbf{Linearity type-checking}};

\node[error] (linmessage) at (\errorMessageX,\linErrorY) {Error message};

\draw [->] (linerror) edge (lintype);
\draw [->] (lintype) edge (linmessage);


\node[rectangle,draw, align=center, text width=1.8cm] (polerror) at (\errorX,\polErrorY) {\textbf{Quantifier error}};

\node[typing, text width=2.5cm] (poltype) at (\errorCheckX,\polErrorY) {\textbf{Quantifier type-checking}};

\node[error] (polmessage) at (\errorMessageX,\polErrorY) {Error message};

\draw [<->] (linerror) edge (polerror);
\draw [->] (polerror) edge (poltype);
\draw [->] (poltype) edge (polmessage);


\node[typing, text width=2.7cm] (difftype) at (\typeX,\lossY) {\textbf{Differentiability type-checking}};

\node[rectangle,draw] (lossfunc) at (\outputX,\lossY) {\textbf{Loss function}};

\node (endloss) [output] at (\backendX,\lossY) {\textbf{Tensorflow}};

\draw [->] (type) edge node [left]{training} (difftype);
\draw [->] (difftype) edge (lossfunc);
\draw [->] (lossfunc) edge (endloss);


\node[text width=2.5cm, typing] (booltype) at (\typeX,\agdaY) {\textbf{Decidability type-checking}};

\node[rectangle,draw] (agdacode) at (\outputX,\agdaY) {\textbf{ITP code}};

\node (endagda) [output] at (\backendX,\agdaY) {\textbf{Agda}};

\draw [->] (type) edge node [left]{export} (booltype);
\draw [->] (booltype) edge (agdacode);
\draw [->] (agdacode) edge (endagda);

\end{tikzpicture}
  \caption{Overview of the \vehicle{} compiler. The same unification-based type-checker is re-used in five different ways to provide support for the different backends.}
    \label{fig:vehicle-type-checker}
\end{figure}

\subsubsection{The Verification Backend}
\label{sec:verifier-backend}

Given a trained network $f$ and a \vehicle{} specification representing $\solutionProperty$, $u$ and $e$, the purpose of the verifier backend is to determine if $\solutionProperty(u \circ f \circ e)$ holds and if not to find a counter-example. \vehicle{} accomplishes this goal by compiling the specification into a set of satisfiability queries for the Marabou verifier~\cite{katz2019marabou}. 

Figure~\ref{fig:marabou-queries} shows the two queries generated from the specification in Figure~\ref{fig:vehicle-controller-spec}. The queries are equisatifiable to the original specification in the sense that a neural network satisfies the specification if and only if no satisfying assignment to variables~$x_0$ and~$x_1$ can be found for either query.
Collectively, the queries represent the property $\networkProperty$. It is important to note that, unlike the original specification which referenced physical quantities such as speed and distances, the Marabou queries refer solely to quantities in the input/output spaces. As a result, the variables and numeric values in these queries are not directly interpretable in terms of the symbolic model of the environment.

\begin{figure}[t]
    \begin{center}
        \begin{subfigure}[t]{0.45\textwidth}
\begin{lstlisting}[language=Haskell]
x0 >= 0.09375
x0 <= 0.90625
x1 >= 0.09375
x1 <= 0.90625
16.0x0 - 8.0x1 + y0 <= 2.75
\end{lstlisting}
        \end{subfigure}
        \hfill
        \begin{subfigure}[t]{0.45\textwidth}
\begin{lstlisting}[language=Haskell]
x0 >= 0.09375
x0 <= 0.90625
x1 >= 0.09375
x1 <= 0.90625
-16.0x0 + 8.0x1 - y0 <= -5.25
\end{lstlisting}
        \end{subfigure}
        \caption{The two Marabou queries, representing $\networkProperty$, generated by the \vehicle{} compiler from the specification in Figure~\ref{fig:vehicle-controller-spec}. Both queries are implicitly existentially quantified over the variables $x_0$ and $x_1$ which represent the inputs to the neural network, and $y_0$ which represents the output of the network. These queries are not seen by a \vehicle{} user in normal operation.}
        \label{fig:marabou-queries}
    \end{center}
\end{figure}

The compilation algorithm interleaves a normalisation-by-evaluation algorithm~\cite{berger1991inverse} with a procedure for the elimination of the quantified problem-space variables. These variables are eliminated by rewriting them in terms of the variables representing the input and outputs of the networks using Gaussian~\cite{atkinson1991introduction} and Fourier-Motzkin~\cite{dantzig1973fourier} elimination, producing a final representation entirely in the network's input/output space.
Notably, the algorithm supports complex specifications , including those with nested applications of multiple networks and quantifiers embedded within the conditions of `if` statements. Ensuring that this procedure remains computationally efficient, particularly for networks with tens of thousands of inputs and outputs, introduces significant complexity. A full description of the compilation procedure is provided in~\cite{DBLP:journals/corr/abs-2402-01353}.

The output of the compilation process is a tree structure, where each internal node represents a conjunction or disjunction, and the leaves correspond to individual verifier queries.
The tree is then stored on disk as the basis of the ``proof cache".
When verification is requested, \vehicle{} loads the cache and traverses the tree, invoking Marabou on each query to decide if there exists a satisfying assignment for the current network and query.
The outcome of each query is recorded in the cache. If any counterexamples for $\networkProperty$ are found in the input-output space, the compiler lifts them to counterexamples for $\solutionProperty$ in the origin problem/result space. Depending on the output of Marabou for each query, Vehicle can then determine that the statement $\solutionProperty(u \circ f \circ e)$ holds. 

It is important to emphasise that \vehicle{} does not expand the class of verifiable specifications beyond what is supported by Marabou. Instead, it enhances the interpretability of those specifications and streamlines the process of generating verification queries. 
Critically, Marabou only supports linear constraints with existential quantifiers. Therefore if a user writes a \vehicle{} specification involving non-linear constraints or alternating quantifiers \vehicle{} will produce an error. 

In such cases, explaining why a specification is not verifiable is an important usability feature. Some of example specifications exceed 300 lines, making it difficult to identify where non-linearities or alternating quantifiers have been introduced. To provide meaningful error messages, we leverage \vehicle{}'s dependent type-checker and instance resolution mechanism to construct a proof of why the original specification is non-linear or has alternating quantifiers. In particular, we replace all types with meta-variables and then re-type check the specification according to a new set of typing rules for the builtin operations. Because type-checking is constructive, the type-checker produces a proof term that the compiler can use to generate clear, actionable explanations. Full details of this procedure are available in~\cite{DaggittAKKA23}.

\subsubsection{The ITP Backend}
\label{sec:itp-backend}

\begin{figure}[t!]
    \input{agda/agda-output-top.tex}
    \vspace{-2.5em}
\begin{multicols}{2}
    \input{agda/agda-output-left.tex}
    \vfill\null
    \columnbreak
    \input{agda/agda-output-right.tex}
\end{multicols}
    \vspace{-2.5em}
    \input{agda/agda-output-bottom.tex}
    \caption{Agda code by \vehicle{} when exporting the specification in Figure~\ref{fig:vehicle-controller-spec}. Import statements are omitted.}
    \label{fig:agda-output}
\end{figure}

The ITP backend is responsible for exporting the the proof of $\solutionProperty(u \circ f \circ e)$ to an ITP, enabling it to be combined with the proof of Lemma~\ref{eq:systemProperty} in order to establish $\systemProperty(\system(u \circ f \circ e))$ in the ITP. Currently, \vehicle{} supports exporting specifications to Agda, with Rocq support nearing completion.
The Agda code generated for the specification in Figure~\ref{fig:vehicle-controller-spec} is shown in Figure~\ref{fig:agda-output}.
Given that most ITP languages are more expressive than \vehicle{} this translation may appear straightforward, however there are two main challenges.

Firstly, as discussed in Section~\ref{sec:decomposition}, in the general case the network and the proof of correctness cannot be represented directly in Agda itself for performance reasons. Instead, as shown in Figure~\ref{fig:agda-output}, the network is included as a \AgdaKeyword{postulate} and an Agda \AgdaMacro{macro} is used to delegate checking the proof to Vehicle. The new challenge lies in preserving Agda's interactivity while ensuring the integrity of the proof. In particular, the Agda proof should fail if the ONNX network file on disk is changed (for example, if the network is retrained after verification).
At the same time, Agda's often invoke Agda's type-checker several times per minute and re-verifying the proof on every interaction is impractical.
The solution is the proof cache discussed in Section~\ref{sec:verifier-backend}.
Alongside the query tree, \vehicle{} stores the hash and location of all networks and datasets used during verifier query compilation within the proof cache.
When Agda queries the validity of the proof, \vehicle{} uses these hashes to verify the integrity of the verification result without having to invoke Marabou again.
As we discuss further in Section~\ref{sec:future-work}, we are exploring the feasability of replacing these hashes with efficiently checkable proof-certificates in the style of~\cite{DBLP:journals/corr/abs-2405-10611}. 

The second challenge arises because users write specifications as Boolean expressions in \vehicle{}, implicitly assuming that the property is decidable. However, ITPs such as Agda require the specification to be encoded at the type-level (e.g. \AgdaFunction{safe} in Figure~\ref{fig:agda-output} and \AgdaFunction{finalState-onRoad} in Section~\ref{sec:example}), where decidability does not generally hold. Despite this, the  conditional term in an `If` expression must be decidable and the therefore the same term in the \vehicle{} specification can be used at both the type level and the Boolean level in the generated ITP code. To determine how to translate the term, we use the same approach as generating error messages for specificationc containing non-linear or alternating quantifiers described in Section~\ref{sec:verifier-backend}. Namely, we leverage \vehicle{}'s dependently-type checker and instance resolution, replacing all Boolean types with meta-variables and then re-solving according to new typing rules for the Boolean builtin operations.

\subsubsection{The Training Backend}
\label{sec:loss-backend}

The final backend enables users to train a network to satisfy a \vehicle{} specification. Numerous techniques have been proposed for incorporating specifications into the training process~\cite{dash2022review,GiunchigliaSL22}. \vehicle{} implements a method known as \emph{differentiable logic}~(DL)~\cite{FischerBDGZV19,KriekenAH22,SKDSS23}, which converts a Boolean-valued specification into a loss function that penalises deviations from the desired property.
The resulting loss function is differentiable almost everywhere with respect to the network weights, allowing it to be used with standard gradient descent algorithms during training. For a broader discussion on the machine-learning justification for using differentiable logic to generate optimisation objectives, see~\cite{flinkow2025generalisedframeworkpropertydrivenmachine}.

Differentiable logic was chosen for two key reasons: a) its generality - depending on the differentiable logic used, any well-typed \vehicle{} specification can be converted to a corresponding loss function; and b) its flexibility - the specification-based training can either be integrated into standard data-driven or reinforcement-based learning workflows or can be applied as an additional fine-tuning step post-training.

\begin{figure}[t!]
    \begin{subfigure}{\textwidth}
	\input{figures/python-training-code.tex}
	\caption{Python code for using the specification in Figure~\ref{fig:vehicle-controller-spec} to train a model. Lines 3-7 contain the Vehicle-specific code needed to generate the loss function. Lines 10-15 are standard code to train a neural network given an arbitrary loss function.}
    \label{fig:python-training}
	\end{subfigure}
	\\
	\vfill
	\vspace{2em}
    \begin{subfigure}{\textwidth}
	\input{figures/python-output.tex}
	\caption{The implementation of the loss function generated automatically by \vehicle{} in Figure~\ref{fig:python-training}. This code is never normally seen by a \vehicle{} user.} 
    \label{fig:loss-function}
	\end{subfigure}
	\caption{\vehicle{}'s training backend.}
    
    \label{fig:training-backend}
\end{figure}

Concretely, given a \texttt{@property p}, \vehicle{} compiles \texttt{p} into a pure function that takes external resources (i.e. networks, datasets and parameters) as inputs and returns a numeric output representing ``how false'' the property \texttt{p} is.
The exact translation method depends on the chosen differential logic.
Numerous logics have been proposed, and \vehicle{} currently implements DL2~\cite{FischerBDGZV19}, G\"{o}del, Product, {\L}ukasiewicz and Yager logics~\cite{KriekenAH22}.

The compilation process procedes in two steps.
Firstly, the specification must be split into two parts: the constraints on the quantified variables and the constraints on the network's behaviour. Critically, the latter needs to be translated into a real-valued formula using the differentiable logic. To do this, we again use the type-checker and instance resolution in a similar fashion to that described in Sections~\ref{sec:verifier-backend}~\&~\ref{sec:itp-backend}.
Next, the same standard normalisation-by-evaluation algorithm as used by the verification algorithm is interleaved with a procedure for splitting the differentiable and the non-differentiable constraints. The former are translated to numeric operations according to the differentiable logic being used. The latter represents the domain of the quantified variables (i.e. the set of values they are allowed to assume) and therefore are attached to the quantifier node in the AST to allow for efficient sampling as described in~\cite{SKDSS23}.

Figure~\ref{fig:python-training} shows the Python code required to be written by the user to train the network using the specification from Section~\ref{sec:example}, while Figure~\ref{fig:loss-function} shows the code that is generated behind the scenes to implement the loss function.
Note that although training via differentiable logic is a very general technique and has shown to be effective in practice~\cite{flinkow2025generalisedframeworkpropertydrivenmachine,flinkowComparingDifferentiableLogics2025}, it does not guarantee that the network will satisfy the specification after training. How to do so is still an open problem.

\subsubsection{Soundness}
\label{sec:correctness}

Given the complexity of the \vehicle{} system and its diverse backends, it is important to ensure its overall soundness.
To this end, we have developed a formal semantics for the \vehicle{} core language, as well as for the target languages used by both the training and the verifier backends.
Based on this foundation, we have proved not only the soundness of the compilation to the two backends, but also that the loss function and the verifier queries generated are logically equivalent in some general sense. This ensures that the propery being trained for is the same as that which we are verifying.
These proofs have been formalised in Agda, and can be found at~\cite{AtkeyVehicleForm23}. We have not formalised the correctness of the ITP backend.

\section{Related and Future Work}\label{sec:future-work}

Given the wide range of areas that \vehicle{} intersects with -- including formal verification, program synthesis, neural network training, and theorem proving -- there is a substantial body of related work. In this section, we highlight some of the most relevant contributions.

\emph{Programming Language Interfaces for Neural Network Verification.}
The need for more conceptual and robust tools and programming language practices
has recently been flagged as one of the biggest challenges in enabling the future development
of neural network verification~\cite{10.1007/978-3-031-91118-7}. Apart from the embedding gap, 
four problems have been flagged as substantial in~\cite{10.1007/978-3-031-91118-7}: the lack of rigorous semantics of specification languages deployed in neural network verification, formalisation and generation of proof certificates, the implementation gap, and support for property-driven training.
\vehicle{} hopes to contribute to resolving most of them.
Other frameworks that provide neural network specification DSLs similar to that of \vehicle{} include DNNV~\cite{shriver2021dnnv} and CAISAR~\cite{girardsatabin2022caisar}. Unlike Vehicle, much of their focus is on improving interoperability between different ATPs for neural network verification, and they do not solve the problem of the embedding gap or integrate training. 

\emph{Explainability and Specifications in Machine Learning.}
\vehicle{}'s methodology applies in domains where a suitable specification is available. Therefore it is not currently easily applicable to domains such as NLP and Computer Vision which defy clearly defined correctness criteria. See~\cite{CasadioKDKKAR22,Casadio2025} for more detailed discussion of this issue.
\vehicle{} is therefore complementary to work such as~\cite{hsieh2022verifying} and~\cite{puasuareanu2023closed} which obtain formal statistical guarantees about neural networks used as sensors in cyber-physical systems.

\emph{Cyber-Physical System Verification.}
Cyber-Physical Systems (CPS) with machine learning components is an important safety-critical use case for neural network verification.
As shown in our running example, a neural network may be utilized as a feedback controller for some plant model, typically represented as ordinary differential equations (ODEs) or generalizations thereof like hybrid automata. These are known as \emph{neural networks control systems (NNCS)}. 
The annual International Competition on Verifying Continuous and Hybrid Systems (ARCH-COMP) 
held with the Applied Verification for Continuous and Hybrid Systems (ARCH) workshop 
has a category for this problem class, known as the AI and NNCS (AINNCS) category~\cite{lopez2019archcomp-ainncs,johnson2020archcomp-ainncs,johnson2021archcomp-ainncs,lopez2022archcomp-ainncs,lopez2023archcomp-ainncs}.
Several approaches for addressing the NNCS verification problem have been developed, such as implemented within software tools like CORA~\cite{kochdumper2023nfm}, JuliaReach~\cite{Bogomolov2019}, NNV~\cite{tran2020cav-tool,lopez2023nnv}, OVERT~\cite{sidrane2019safeml}, POLAR~\cite{huang2022atva}, Sherlock~\cite{dutta2018adhs,dutta2019hscc}, ReachNN*~\cite{huang2019reachnn,fan2020reachnn}, VenMAS~\cite{AkintundeBKL20}, and Verisig~\cite{ivanov2019hscc,ivanov2020tecs,ivanov2021cav}.
More broadly, researchers have considered several strategies for the specification of properties of CPS with neural network components~\cite{fremont2019pldi,astorga2023oopsla,calinescu2024tse}.
These cover significant challenges in the CPS domain, ranging from classical software verification problems to real-time systems concerns, scalability, as well as finding suitable specifications~\cite{seshia2022cacm,tran2022mdat,DBLP:journals/corr/abs-2402-10998,DBLP:journals/corr/abs-2405-14058}. Crucially however, many of these techniques can not be easily linked to more general purpose tools such as ITPs for reasoning about the environments. We believe this area would also benefit from a more principled programming language support, and languages like \vehicle{} can provide a trustworthy infrastructure for consistent specifications of these complex systems. 

\emph{Neuro-Symbolic Programs and Proof-Carrying Code.}
Finally, our work on \vehicle{} also relates to the nascent field of \emph{neuro-symbolic programming}, seen as a collection of methods of merging machine learning code and standard (symbolic) code~\cite{PGL-049}. In particular, \vehicle{} can be seen as a step towards the goal of enabling a \emph{proof-carrying neuro-symbolic code}~\cite{komendantskaya2025proofcarryingneurosymboliccode}. The idea is to use a combination of formal methods and compilation techniques 
to enable  light-weight verification of complex neuro-symbolic interfaces in the style of \emph{self-certifying code}~\cite{10.1145/3689624}.  

Our priorities for improving Vehicle include:
\begin{itemize}
\item \textbf{Proof Certificates}. Currently the ITP must trust Vehicle's assertion that the network satisfies the specification. As discussed in Section~\ref{subsec:egap}, directly representing large neural networks within the ITP is likely infeasible. However, we are exploring ways to adapt Vehicle and the ATPs to generate proof certificates that can be efficiently checked by the ITP itself.
The feasibility of checking ATP certificates was demonstrated in~\cite{DBLP:journals/corr/abs-2405-10611}. 
\item \textbf{ITP Backends}. \vehicle{} was designed with a view of providing a principled (and sound) way of interfacing to many ITPs,
depending on the demands of the `symbolic' component verification. For example, industrial ITPs such as Imandra offer stronger automation and libraries that support infinite-precision reals as well as floats; Rocq has extensive Measure theory libraries~\cite{DBLP:journals/jar/AffeldtC23,affeldt2022mathcomp}; KeYmaera X~\cite{fulton2015keymaeraX} is designed for reasoning about cyber-physical systems with continuous dynamics. Such features may facilitate future CPS verification projects, where \vehicle{} can help to interface with neural network verifiers.   
Recent Rocq backend for Vehicle has proven easy to implement~\cite{Smart25}, and has demonstrated  \vehicle{}'s readiness for future ITP extensions. 
\item \textbf{Numeric Quantisation}. Currently the \vehicle{} syntax and semantics assume the neural networks operate over real numbers. However, in practice neural networks are implemented using quantised floating point values, with a precision of anywhere between 4 and 32 bits. This mismatch in the semantics has been shown to affect the soundness of the neural network verifiers themselves~\cite{jia2021exploiting,10.1007/978-3-031-91118-7}. How best to address the quantisation issue during verification is an open problem.
\end{itemize}


\section{Conclusions}

In this paper we have identified the embedding gap as an existing problem in the verification of neural-symbolic programs and described \vehicle{}, the first tool that aims to bridge that gap.
We have shown how \vehicle{} facilitates proofs about the correctness of neuro-symbolic programs by linking specifications to training frameworks, verifiers and ITPs.
We have also demonstrated its utility by verifying the correctness of a neuro-symbolic car controller.
We believe this to be the first ever modular proof of the complete verification of a neuro-symbolic program that utilises both ATPs and ITPs.

Our example is, of course, a toy scenario that was primarily chosen because it is small enough to fit in this paper. In a real-world scenario, the environmental dynamics are far more complicated and the car controller will have other objectives such as reaching way points and obstacle avoidance. Therefore we believe one of the overall challenges in this field is to work out how to construct our neuro-symbolic systems so that the safety critical properties (i.e. staying on the road, collision avoidance) are formally verifiable, while allowing the neural components to optimise for the non-safety-critical goals.

\section{Contribution statement}

Conceptualisation and analysis by Daggitt, Kokke, Atkey and Komendantskaya with help from Arnaboldi. Implementation of \vehicle{} by Daggitt and Kokke with help from Atkey and Slursarz. Manuscript preparation by Daggitt, Komendantskaya and Atkey.

\bibliographystyle{splncs04}
\bibliography{bibliography}

\end{document}

%% file: agda-unicode.tex
\usepackage{newunicodechar}

\newunicodechar{𝕋}{\ensuremath{\mathbb{T}}}
\newunicodechar{ℚ}{\ensuremath{\mathbb{Q}}}
\newunicodechar{ℤ}{\ensuremath{\mathbb{Z}}}
\newunicodechar{ℕ}{\ensuremath{\mathbb{N}}}
\newunicodechar{ℓ}{\ensuremath{\ell}}
\newunicodechar{λ}{\ensuremath{\lambda}}
\newunicodechar{∞}{\ensuremath{\infty}}

\newunicodechar{∣}{\ensuremath{\mid}}
\newunicodechar{⊔}{\ensuremath{\sqcup}}
\newunicodechar{∸}{\ensuremath{\dotdiv}}
\newunicodechar{∥}{\ensuremath{\parallel}}

\newunicodechar{≤}{\ensuremath{\leq}}
\newunicodechar{≱}{\ensuremath{\ngeq}}
\newunicodechar{▷}{\ensuremath{\vartriangleright}}
\newunicodechar{⇿}{\ensuremath{\leftrightarrow}} 
\newunicodechar{∷}{\ensuremath{::}}

\newunicodechar{⌊}{\ensuremath{\lfloor}}
\newunicodechar{⌋}{\ensuremath{\rfloor}}
\newunicodechar{∨}{\ensuremath{\vee}}

\newunicodechar{≢}{\ensuremath{\nequiv}}
\newunicodechar{≈}{\ensuremath{\approx}}
\newunicodechar{≉}{\ensuremath{\napprox}}
\newunicodechar{≟}{\ensuremath{\stackrel{?}{=}}}

\newunicodechar{∀}{\ensuremath{\forall}}
\newunicodechar{∃}{\ensuremath{\exists}}
\newunicodechar{⇒}{\ensuremath{\Rightarrow}}
\newunicodechar{∧}{\ensuremath{\wedge}}
\newunicodechar{∈}{\ensuremath{\in}}
\newunicodechar{∉}{\ensuremath{\notin}}
\newunicodechar{∴}{\ensuremath{\therefore}}
\newunicodechar{∎}{\ensuremath{\qed}}
\newunicodechar{∘}{\ensuremath{\circ}}
\newunicodechar{⊕}{\ensuremath{\oplus}}

\newunicodechar{ₐ}{\ensuremath{_a}}
\newunicodechar{ₑ}{\ensuremath{_e}}
\newunicodechar{ᵢ}{\ensuremath{_i}}
\newunicodechar{ₖ}{\ensuremath{_k}}
\newunicodechar{ₗ}{\ensuremath{_l}}
\newunicodechar{ₘ}{\ensuremath{_m}}
\newunicodechar{ₙ}{\ensuremath{_n}}
\newunicodechar{ₓ}{\ensuremath{_x}}

\newunicodechar{₀}{\ensuremath{_0}}
\newunicodechar{₁}{\ensuremath{_1}}
\newunicodechar{₂}{\ensuremath{_2}}
\newunicodechar{₃}{\ensuremath{_3}}
\newunicodechar{₊}{\ensuremath{_+}}

\newunicodechar{ᵉ}{\ensuremath{^e}}
\newunicodechar{ᵐ}{\ensuremath{^m}}
\newunicodechar{ʳ}{\ensuremath{^r}}

\newunicodechar{∅}{\ensuremath{\emptyset}}
\newunicodechar{⊥}{\ensuremath{\bot}}

%% file: macros.tex
\newcommand{\nat}{\mathbb{N}}
\newcommand{\real}{\mathbb{R}}

\newcommand{\cons}[2]{#1 :: #2}

\newcommand{\bop}[3]{\ensuremath{#1 \: #2 \: #3}}

\newcommand{\hypothesis}{\mathcal{H}}
\newcommand{\problemSpace}{\mathcal{P}}
\newcommand{\resultSpace}{\mathcal{R}}

\newcommand{\system}{s}
\newcommand{\systemProperty}{\Psi}
\newcommand{\solutionProperty}{\Phi}
\newcommand{\networkProperty}{\Xi}

\newcommand{\VSL}{\textsc{VCL}}

\newcommand{\grammarItem}[1]{\langle \textit{#1} \, \rangle}

\newcommand{\identClass}{\grammarItem{id}}

\newcommand{\piSymbol}{\Pi}

\newcommand{\boolSymbol}{\texttt{Bool}}
\newcommand{\realSymbol}{\texttt{Real}}
\newcommand{\vecSymbol}{\texttt{Tensor}}
\newcommand{\indexSymbol}{\texttt{Index}}

\newcommand{\typeSymbol}{\texttt{Type}}

\newcommand{\atSymbol}{\texttt{!}}
\newcommand{\andSymbol}{\texttt{and}}
\newcommand{\orSymbol}{\texttt{or}}

\newcommand{\notSymbol}{\texttt{not}}
\newcommand{\addSymbol}{\texttt{+}}
\newcommand{\mulSymbol}{\texttt{*}}
\newcommand{\eqSymbol}{\texttt{==}}
\newcommand{\neqSymbol}{\texttt{!=}}
\newcommand{\leqSymbol}{\texttt{<=}}
\newcommand{\leSymbol}{\texttt{<}}
\newcommand{\forallSymbol}{\texttt{forall}}
\newcommand{\existsSymbol}{\texttt{exists}}

\newcommand{\trueSymbol}{\texttt{true}}
\newcommand{\falseSymbol}{\texttt{false}}
\newcommand{\foreachSymbol}{\texttt{foreach}}
\newcommand{\foldSymbol}{\texttt{foldr}}

\newcommand{\gprog}{\ensuremath{p}}

\newcommand{\gexpr}{\ensuremath{e}}
\newcommand{\gvar}{\ensuremath{x}}

\newcommand{\hDef}[3]{\ensuremath{\texttt{function}\: #1 : #2 = #3}}

\newcommand{\hNetworkDecl}[2]{\ensuremath{\texttt{@network}\: #1 : #2}}
\newcommand{\hDatasetDecl}[2]{\ensuremath{\texttt{@dataset}\: #1 : #2}}
\newcommand{\hParameterDecl}[2]{\ensuremath{\texttt{@parameter}\: #1 : #2}}
\newcommand{\hPropertyDecl}[3]{\ensuremath{\texttt{@property}\ #1 : #2 = #3}}

\newcommand{\hBinder}[2]{(\hVar{#1} : #2)}

\newcommand{\hLam}[3]{\ensuremath{\lambda \: \hBinder{#1}{#2} \texttt{.} #3}}
\newcommand{\hApp}[2]{\ensuremath{#1 \: #2}}

\newcommand{\hPi}[3]{\ensuremath{\piSymbol \hBinder{#1}{#2} \texttt{.} #3}}

\newcommand{\hVar}[1]{\ensuremath{\texttt{#1}}}

\newcommand{\hType}{\typeSymbol}

\newcommand{\hRealType}{\ensuremath{\realSymbol{}}}

\newcommand{\hAdd}[2]{\bop{#1}{\addSymbol{}}{#2}}
\newcommand{\hMul}[2]{\bop{#1}{\mulSymbol{}}{#2}}

\newcommand{\hBoolType}{\ensuremath{\boolSymbol{}}}

\newcommand{\hTrue}{\trueSymbol}
\newcommand{\hFalse}{\falseSymbol}
\newcommand{\hNot}[1]{\ensuremath{\notSymbol{} \: #1}}
\newcommand{\hAnd}[2]{\bop{#1}{\andSymbol{}}{#2}}
\newcommand{\hOr}[2]{\bop{#1}{\orSymbol{}}{#2}}

\newcommand{\hEq}[2]{\bop{#1}{\eqSymbol{}}{#2}}
\newcommand{\hNeq}[2]{\bop{#1}{\neqSymbol{}}{#2}}
\newcommand{\hLeq}[2]{\bop{#1}{\leqSymbol{}}{#2}}
\newcommand{\hLe}[2]{\bop{#1}{\leSymbol{}}{#2}}
\newcommand{\hIf}[3]{\ensuremath{\texttt{if} \:\: #1 \:\: \texttt{then} \:\: #2 \:\: \texttt{else} \:\: #3}}
\newcommand{\hForall}[3]{\ensuremath{\forallSymbol{} \: \hBinder{#1}{#2} \texttt{.} \: #3}}
\newcommand{\hExists}[3]{\ensuremath{\existsSymbol{} \: \hBinder{#1}{#2} \texttt{.} \: #3}}
\newcommand{\hFold}[3]{\ensuremath{\hApp{\hApp{\hApp{\foldSymbol{}}{#1}}{#2}}{#3}}}

\newcommand{\hVecType}[2]{\ensuremath{\vecSymbol{} \: #1 \: #2}}
\newcommand{\hIndexType}[1]{\ensuremath{\indexSymbol{} \: #1}}
\newcommand{\hAt}[2]{\bop{#1}{\atSymbol{}}{#2}}
\newcommand{\hSeq}[1]{\ensuremath{[#1]}}

\newcommand{\hForeach}[3]{\foreachSymbol{} \hBinder{#1}{#2} \texttt{.} #3}

\newcommand{\vehicle}{\textsc{Vehicle}}



%% file: figures/specification-diagram.tex
\begin{tikzpicture}[thick, scale=0.7, font=\sffamily\footnotesize,
    set/.style = {circle,
        maximum size = 2cm}, -> /.tip = Stealth]

\newcommand{\rowOneY}{11.8};
\newcommand{\rowTwoY}{9.3};
\newcommand{\colFourX}{15.5};

\filldraw [color=blue!30] (7,14) rectangle ++ (-7,-6);

\filldraw [color=red!30] (17.5,14) rectangle ++ (-7.5,-6);

\path [<->,dashed] (7, 12.5) edge (10, 12.5);

\node[align=center, fill=white, inner sep=0, outer sep=0] at (8.5,13.2) {Embedding \\ gap};

\node[align=center] at (3.5,13.5) {\underline{Problem space}};

\node[align=center] at (13.5,13.5) {\underline{Embedding space}};

\node[rectangle, text width=1.8cm, draw, align=center] (A) at (5,\rowOneY) {Specification of $\solutionProperty$, $e$, $u$};
\node[align=center] at (2,\rowOneY) {Specification \\ language};

\node[rectangle, text width=1.8cm, draw, align=center] (B) at (12,\rowOneY) {Training with $\networkProperty$};
\node[align=center] at (\colFourX,\rowOneY) {\underline{Training platform} \\ Tensorflow etc. };

\draw [->] (A) edge (B);

\node[rectangle,draw,text width=1.8cm, align=center] (D) at (12,\rowTwoY) {Verification of $\networkProperty$};
\node[text width=2cm, align=center] at (\colFourX,\rowTwoY) {\underline{ATPs} \\ Marabou etc.};

\draw [->] (A) edge (D);

\node[rectangle,draw,text width=1.8cm, align=center] (E) at (5,\rowTwoY) {Integration of $\solutionProperty$ with $\systemProperty$};
\node[text width=1.7cm, align=center] at (2,\rowTwoY) {\underline{ITPs} \\ Agda etc.};

\draw [->] (A) edge (E);
\draw [->, dashed] (D) edge (E);
\draw [->, dashed] (B) edge (D);

\draw [->, dashed] (D) edge[bend left=30] (B);
\end{tikzpicture}

%% file: agda/state.tex
\begin{code}%
\>[0]\AgdaKeyword{record}\AgdaSpace{}%
\AgdaRecord{State}\AgdaSpace{}%
\AgdaSymbol{:}\AgdaSpace{}%
\AgdaPrimitive{Set}\AgdaSpace{}%
\AgdaKeyword{where}\<%
\\
\>[0][@{}l@{\AgdaIndent{0}}]%
\>[2]\AgdaKeyword{constructor}\AgdaSpace{}%
\AgdaInductiveConstructor{state}\<%
\\
\>[2]\AgdaKeyword{field}\<%
\\
\>[2][@{}l@{\AgdaIndent{0}}]%
\>[4]\AgdaField{windSpeed}\AgdaSpace{}%
\>[14]\AgdaSymbol{:}\AgdaSpace{}%
\AgdaRecord{ℚ}\<%
\\
\>[4]\AgdaField{position}%
\>[14]\AgdaSymbol{:}\AgdaSpace{}%
\AgdaRecord{ℚ}\<%
\\
\>[4]\AgdaField{velocity}%
\>[14]\AgdaSymbol{:}\AgdaSpace{}%
\AgdaRecord{ℚ}\<%
\\
\>[4]\AgdaField{sensor}%
\>[14]\AgdaSymbol{:}\AgdaSpace{}%
\AgdaRecord{ℚ}\<%
\end{code}

%% file: agda/observation.tex
\begin{code}%
\>[0]\AgdaKeyword{record}\AgdaSpace{}%
\AgdaRecord{Observation}\AgdaSpace{}%
\AgdaSymbol{:}\AgdaSpace{}%
\AgdaPrimitive{Set}\AgdaSpace{}%
\AgdaKeyword{where}\<%
\\
\>[0][@{}l@{\AgdaIndent{0}}]%
\>[2]\AgdaKeyword{constructor}\AgdaSpace{}%
\AgdaInductiveConstructor{observe}\<%
\\
\>[2]\AgdaKeyword{field}\<%
\\
\>[2][@{}l@{\AgdaIndent{0}}]%
\>[4]\AgdaField{windShift}%
\>[16]\AgdaSymbol{:}\AgdaSpace{}%
\AgdaRecord{ℚ}\<%
\\
\>[4]\AgdaField{sensorError}\AgdaSpace{}%
\>[16]\AgdaSymbol{:}\AgdaSpace{}%
\AgdaRecord{ℚ}\<%
\end{code}

%% file: agda/controller-type.tex
\begin{code}%
\>[0]\AgdaFunction{controller}\AgdaSpace{}%
\AgdaSymbol{:}\AgdaSpace{}%
\AgdaRecord{ℚ}\AgdaSpace{}%
\AgdaSymbol{→}\AgdaSpace{}%
\AgdaRecord{ℚ}\AgdaSpace{}%
\AgdaSymbol{→}\AgdaSpace{}%
\AgdaRecord{ℚ}\<%
\end{code}

%% file: agda/nextState.tex
\begin{code}%
\>[0]\AgdaFunction{nextState}\AgdaSpace{}%
\AgdaSymbol{:}\AgdaSpace{}%
\AgdaRecord{Observation}\AgdaSpace{}%
\AgdaSymbol{→}\AgdaSpace{}%
\AgdaRecord{State}\AgdaSpace{}%
\AgdaSymbol{→}\AgdaSpace{}%
\AgdaRecord{State}\<%
\\
\>[0]\AgdaFunction{nextState}\AgdaSpace{}%
\AgdaBound{o}\AgdaSpace{}%
\AgdaBound{s}\AgdaSpace{}%
\AgdaSymbol{=}\AgdaSpace{}%
\AgdaInductiveConstructor{state}\AgdaSpace{}%
\AgdaFunction{newWindSpeed}\AgdaSpace{}%
\AgdaFunction{newPosition}\AgdaSpace{}%
\AgdaFunction{newVelocity}\AgdaSpace{}%
\AgdaFunction{newSensor}\<%
\\
\>[0][@{}l@{\AgdaIndent{0}}]%
\>[2]\AgdaKeyword{where}\<%
\\
\>[2]\AgdaFunction{newWindSpeed}\AgdaSpace{}%
\>[15]\AgdaSymbol{=}\AgdaSpace{}%
\AgdaField{windSpeed}\AgdaSpace{}%
\AgdaBound{s}\AgdaSpace{}%
\AgdaOperator{\AgdaFunction{+}}\AgdaSpace{}%
\AgdaField{windShift}\AgdaSpace{}%
\AgdaBound{o}\<%
\\
\>[2]\AgdaFunction{newPosition}%
\>[15]\AgdaSymbol{=}\AgdaSpace{}%
\AgdaField{position}\AgdaSpace{}%
\AgdaBound{s}%
\>[29]\AgdaOperator{\AgdaFunction{+}}\AgdaSpace{}%
\AgdaField{velocity}\AgdaSpace{}%
\AgdaBound{s}\AgdaSpace{}%
\AgdaOperator{\AgdaFunction{+}}\AgdaSpace{}%
\AgdaFunction{newWindSpeed}\<%
\\
\>[2]\AgdaFunction{newSensor}%
\>[15]\AgdaSymbol{=}\AgdaSpace{}%
\AgdaFunction{newPosition}\AgdaSpace{}%
\AgdaOperator{\AgdaFunction{+}}\AgdaSpace{}%
\AgdaField{sensorError}\AgdaSpace{}%
\AgdaBound{o}\<%
\\
\>[2]\AgdaFunction{newVelocity}%
\>[15]\AgdaSymbol{=}\AgdaSpace{}%
\AgdaField{velocity}\AgdaSpace{}%
\AgdaBound{s}%
\>[29]\AgdaOperator{\AgdaFunction{+}}\AgdaSpace{}%
\AgdaFunction{controller}\AgdaSpace{}%
\AgdaFunction{newSensor}\AgdaSpace{}%
\AgdaSymbol{(}\AgdaField{sensor}\AgdaSpace{}%
\AgdaBound{s}\AgdaSymbol{)}\<%
\end{code}

%% file: agda/finalState.tex
\begin{code}%
\>[0]\AgdaFunction{finalState}\AgdaSpace{}%
\AgdaSymbol{:}\AgdaSpace{}%
\AgdaDatatype{List}\AgdaSpace{}%
\AgdaRecord{Observation}\AgdaSpace{}%
\AgdaSymbol{→}\AgdaSpace{}%
\AgdaRecord{State}\<%
\\
\>[0]\AgdaFunction{finalState}\AgdaSpace{}%
\AgdaBound{xs}\AgdaSpace{}%
\AgdaSymbol{=}\AgdaSpace{}%
\AgdaFunction{foldr}\AgdaSpace{}%
\AgdaFunction{nextState}\AgdaSpace{}%
\AgdaFunction{initialState}\AgdaSpace{}%
\AgdaBound{xs}\<%
\end{code}

%% file: agda/finalState-onRoad.tex
\begin{code}%
\>[0]\AgdaFunction{finalState-onRoad}\AgdaSpace{}%
\AgdaSymbol{:}\AgdaSpace{}%
\AgdaSymbol{∀}\AgdaSpace{}%
\AgdaBound{xs}\AgdaSpace{}%
\AgdaSymbol{→}\AgdaSpace{}%
\AgdaDatatype{All}\AgdaSpace{}%
\AgdaFunction{ValidObservation}\AgdaSpace{}%
\AgdaBound{xs}\AgdaSpace{}%
\AgdaSymbol{→}\AgdaSpace{}%
\AgdaFunction{OnRoad}\AgdaSpace{}%
\AgdaSymbol{(}\AgdaFunction{finalState}\AgdaSpace{}%
\AgdaBound{xs}\AgdaSymbol{)}\<%
\end{code}

%% file: agda/controller-lemma.tex
\begin{code}%
\>[0]\AgdaFunction{controller-lemma}\AgdaSpace{}%
\>[10]\AgdaSymbol{:}%
\>[332I]\AgdaSymbol{∀}\AgdaSpace{}%
\AgdaBound{x}\AgdaSpace{}%
\AgdaBound{y}\AgdaSpace{}%
\AgdaSymbol{→}\AgdaSpace{}%
\AgdaOperator{\AgdaFunction{∣}}\AgdaSpace{}%
\AgdaBound{x}\AgdaSpace{}%
\AgdaOperator{\AgdaFunction{∣}}\AgdaSpace{}%
\AgdaOperator{\AgdaDatatype{≤}}\AgdaSpace{}%
\AgdaBound{3.25}\AgdaSpace{}%
\AgdaSymbol{→}\AgdaSpace{}%
\AgdaOperator{\AgdaFunction{∣}}\AgdaSpace{}%
\AgdaBound{y}\AgdaSpace{}%
\AgdaOperator{\AgdaFunction{∣}}\AgdaSpace{}%
\AgdaOperator{\AgdaDatatype{≤}}\AgdaSpace{}%
\AgdaBound{3.25}\AgdaSpace{}%
\AgdaSymbol{→}\AgdaSpace{}%
\AgdaOperator{\AgdaFunction{∣}}\AgdaSpace{}%
\AgdaFunction{controller}\AgdaSpace{}%
\AgdaBound{x}\AgdaSpace{}%
\AgdaBound{y}\AgdaSpace{}%
\AgdaOperator{\AgdaFunction{+}}\AgdaSpace{}%
\AgdaFunction{2}\AgdaSpace{}%
\AgdaOperator{\AgdaFunction{*}}\AgdaSpace{}%
\AgdaBound{x}\AgdaSpace{}%
\AgdaOperator{\AgdaFunction{-}}\AgdaSpace{}%
\AgdaBound{y}\AgdaSpace{}%
\AgdaOperator{\AgdaFunction{∣}}\AgdaSpace{}%
\AgdaOperator{\AgdaDatatype{<}}\AgdaSpace{}%
\AgdaBound{1.25}\AgdaSpace{}\<%
\end{code}

%% file: figures/vehicle-controller-spec.tex
\begin{minted}[autogobble]{haskell}
    type Input = Tensor Rat [2]
    currentPosition = 0
    previousSensor = 1
    
    type Output = Tensor Rat [1]
    velocity = 0
    
    @network 
    controller : Input -> Output
    
    normalise : Input -> Input
    normalise x = forall i . (x ! i + 4.0) / 8.0
    
    safeInput : Input -> Bool
    safeInput x = -3.25 <= x ! currentSensor <= 3.25 and 
                  -3.25 <= x ! previousSensor <= 3.25
    
    safeOutput : Output -> Bool
    safeOutput x = let y = controller (normalise x) ! velocity in
        -1.25 < y + 2 * x ! currentSensor - x ! previousSensor < 1.25
    
    @property
    safe : Bool
    safe = forall x . safeInput x => safeOutput x        
    \end{minted}

%% file: figures/vehicle-syntax.tex
\centering
\begin{subfigure}[t]{0.32\textwidth}
\begin{grammar}
<prog> $\ni p$ ::= [  ] | \cons{d}{\gprog}

<decl> $\ni d$ ::= \phantom{}
	\alt \hDef{\identClass}{\gexpr}{\gexpr}
	\alt \hNetworkDecl{\identClass}{\gexpr}
	\alt \hDatasetDecl{\identClass}{\gexpr}
	\alt \hParameterDecl{\identClass}{\gexpr}
	\alt \hPropertyDecl{\identClass}{\gexpr}{\gexpr}

\end{grammar}
\end{subfigure}
\hfill
\begin{subfigure}[t]{0.67\textwidth}
\begin{grammar}
<expr> $\ni e$ ::=  \\
	\hType{}
	| $\hPi{\gvar}{\gexpr}{\gexpr}$
	| $\hLam{\gvar}{\gexpr}{\gexpr}$
	| $\gvar$ 
	| $\hApp{\gexpr}{\gexpr}$
	| 
	\\
	\hBoolType{}
	| \hTrue{}
	| \hFalse{}
	| \hNot{\gexpr}
	| \hAnd{\gexpr}{\gexpr}
	| \hOr{\gexpr}{\gexpr} 
	|
	\\
	\hForall{\gvar}{\gexpr}{\gexpr}
	| \hExists{\gvar}{\gexpr}{\gexpr}
	|
	\\
	\hIf{\gexpr}{\gexpr}{\gexpr}
	|
	\\
	\hRealType{}
	| $r \in \real$ 
	| \hAdd{\gexpr}{\gexpr}  
	| \hMul{\gexpr}{\gexpr}  
	| \hEq{\gexpr}{\gexpr}
	| \hNeq{\gexpr}{\gexpr} 
	| \hLeq{\gexpr}{\gexpr}
	| \hLe{\gexpr}{\gexpr}
	|
	\\
	\hVecType{\gexpr}{\gexpr}
	| \hIndexType{\gexpr}
	| $n \in \nat$
	| \hSeq{\gexpr, $\ldots$, \gexpr}
	| \hAt{\gexpr}{\gexpr}
	| 
	\\
	\hForeach{\gvar}{n}{\gexpr}
	| \hFold{\gexpr}{\gexpr}{\gexpr}
\end{grammar}
\end{subfigure}

%% file: agda/agda-output-top.tex
\begin{code}%
\>[0]\AgdaKeyword{module}\AgdaSpace{}%
\AgdaFunction{ControllerSpecification}\AgdaSpace{}\AgdaKeyword{where}\<%
\end{code}

%% file: agda/agda-output-left.tex
\begin{code}%
\>[0]\AgdaFunction{InputVector}\AgdaSpace{}%
\AgdaSymbol{:}\AgdaSpace{}%
\AgdaPrimitive{Set}\<%
\\
\>[0]\AgdaFunction{InputVector}\AgdaSpace{}%
\AgdaSymbol{=}\AgdaSpace{}%
\AgdaFunction{Tensor}\AgdaSpace{}%
\AgdaRecord{ℚ}\AgdaSpace{}%
\AgdaSymbol{(}\AgdaNumber{2}\AgdaSpace{}%
\AgdaOperator{\AgdaInductiveConstructor{∷}}\AgdaSpace{}%
\AgdaInductiveConstructor{[]}\AgdaSymbol{)}\<%
\\
\\
\>[0]\AgdaFunction{currentSensor}\AgdaSpace{}%
\AgdaSymbol{:}\AgdaSpace{}%
\AgdaFunction{Fin}\AgdaSpace{}%
\AgdaNumber{2}\<%
\\
\>[0]\AgdaFunction{currentSensor}\AgdaSpace{}%
\AgdaSymbol{=}\AgdaSpace{}%
\AgdaOperator{\AgdaFunction{\#}}\AgdaSpace{}%
\AgdaNumber{0}\<%
\\
\\
\>[0]\AgdaFunction{previousSensor}\AgdaSpace{}%
\AgdaSymbol{:}\AgdaSpace{}%
\AgdaFunction{Fin}\AgdaSpace{}%
\AgdaNumber{2}\<%
\\
\>[0]\AgdaFunction{previousSensor}\AgdaSpace{}%
\AgdaSymbol{=}\AgdaSpace{}%
\AgdaOperator{\AgdaFunction{\#}}\AgdaSpace{}%
\AgdaNumber{1}\<%
\end{code}

%% file: agda/agda-output-right.tex
\begin{code}%
\>[0]\AgdaFunction{OutputVector}\AgdaSpace{}%
\AgdaSymbol{:}\AgdaSpace{}%
\AgdaPrimitive{Set}\<%
\\
\>[0]\AgdaFunction{OutputVector}\AgdaSpace{}%
\AgdaSymbol{=}\AgdaSpace{}%
\AgdaFunction{Tensor}\AgdaSpace{}%
\AgdaRecord{ℚ}\AgdaSpace{}%
\AgdaSymbol{(}\AgdaNumber{1}\AgdaSpace{}%
\AgdaOperator{\AgdaInductiveConstructor{∷}}\AgdaSpace{}%
\AgdaInductiveConstructor{[]}\AgdaSymbol{)}\<%
\\
\\
\>[0]\AgdaFunction{velocity}\AgdaSpace{}%
\AgdaSymbol{:}\AgdaSpace{}%
\AgdaFunction{Fin}\AgdaSpace{}%
\AgdaNumber{1}\<%
\\
\>[0]\AgdaFunction{velocity}\AgdaSpace{}%
\AgdaSymbol{=}\AgdaSpace{}%
\AgdaOperator{\AgdaFunction{\#}}\AgdaSpace{}%
\AgdaNumber{0}\<%
\end{code}

%% file: agda/agda-output-bottom.tex
\begin{code}%
\>[0]\AgdaKeyword{postulate}\AgdaSpace{}\AgdaFunction{controller}\AgdaSpace{}%
\AgdaSymbol{:}\AgdaSpace{}%
\AgdaFunction{InputVector}\AgdaSpace{}%
\AgdaSymbol{→}\AgdaSpace{}%
\AgdaRecord{OutputVector}\<%
\\
\\
\>[0]\AgdaFunction{normalise}\AgdaSpace{}%
\AgdaSymbol{:}\AgdaSpace{}%
\AgdaFunction{InputVector}\AgdaSpace{}%
\AgdaSymbol{→}\AgdaSpace{}%
\AgdaFunction{InputVector}\<%
\\
\>[0]\AgdaFunction{normalise}\AgdaSpace{}%
\AgdaBound{x}\AgdaSpace{}%
\AgdaSymbol{=}\AgdaSpace{}%
\AgdaSymbol{λ}\AgdaSpace{}%
\AgdaBound{i}\AgdaSpace{}%
\AgdaSymbol{→}\AgdaSpace{}%
\AgdaSymbol{(}\AgdaBound{x}\AgdaSpace{}%
\AgdaBound{i}\AgdaSpace{}%
\AgdaOperator{\AgdaFunction{ℚ.+}}\AgdaSpace{}%
\AgdaNumber{4}\AgdaSymbol{)}\AgdaSpace{}%
\AgdaOperator{\AgdaFunction{ℚ.÷}}\AgdaSpace{}%
\AgdaNumber{8}\<%
\\
\\
\>[0]\AgdaFunction{SafeInput}\AgdaSpace{}%
\AgdaSymbol{:}\AgdaSpace{}%
\AgdaFunction{InputVector}\AgdaSpace{}%
\AgdaSymbol{→}\AgdaSpace{}%
\AgdaPrimitive{Set}\<%
\\
\>[0]\AgdaFunction{SafeInput}\AgdaSpace{}%
\AgdaBound{x}\AgdaSpace{}%
\>[12]\AgdaSymbol{=}\AgdaSpace{}%
\AgdaSymbol{(}\AgdaOperator{\AgdaFunction{ℚ.-}}\AgdaSpace{}%
\AgdaNumber{3.25}\AgdaSpace{}%
\AgdaOperator{\AgdaDatatype{ℚ.≤}}\AgdaSpace{}%
\AgdaFunction{currentSensor}\AgdaSpace{}%
\AgdaBound{x}\AgdaSpace{}%
\AgdaOperator{\AgdaFunction{×}}\AgdaSpace{}%
\AgdaFunction{currentSensor}\AgdaSpace{}%
\AgdaBound{x}\AgdaSpace{}%
\AgdaOperator{\AgdaDatatype{ℚ.≤}}\AgdaSpace{}%
\AgdaNumber{3.25}\AgdaSymbol{)}\AgdaSpace{}\<%
\\
\>[12]\AgdaOperator{\AgdaFunction{×}}\AgdaSpace{}%
\AgdaSymbol{(}\AgdaOperator{\AgdaFunction{ℚ.-}}\AgdaSpace{}%
\AgdaNumber{3.25}\AgdaSpace{}%
\AgdaOperator{\AgdaDatatype{ℚ.≤}}\AgdaSpace{}%
\AgdaFunction{previousSensor}\AgdaSpace{}%
\AgdaBound{x}\AgdaSpace{}%
\AgdaOperator{\AgdaFunction{×}}\AgdaSpace{}%
\AgdaFunction{previousSensor}\AgdaSpace{}%
\AgdaBound{x}\AgdaSpace{}%
\AgdaOperator{\AgdaDatatype{ℚ.≤}}\AgdaSpace{}%
\AgdaNumber{3.25}\AgdaSymbol{)}\<%
\\
\\
\>[0]\AgdaFunction{SafeOutput}\AgdaSpace{}%
\AgdaSymbol{:}\AgdaSpace{}%
\AgdaFunction{InputVector}\AgdaSpace{}%
\AgdaSymbol{→}\AgdaSpace{}%
\AgdaPrimitive{Set}\<%
\\
\>[0]\AgdaFunction{SafeOutput}\AgdaSpace{}%
\AgdaBound{x}\AgdaSpace{}%
\>[12]\AgdaSymbol{=}\AgdaSpace{}\<%
\\
\>[0][@{}l@{\AgdaIndent{0}}]%
\>[2]\AgdaKeyword{let}\AgdaSpace{}%
\AgdaBound{y}\AgdaSpace{}%
\AgdaSymbol{=}\AgdaSpace{}%
\AgdaPostulate{controller}\AgdaSpace{}%
\AgdaSymbol{(}\AgdaFunction{normalise}\AgdaSpace{}%
\AgdaBound{x}\AgdaSymbol{)}\AgdaSpace{}%
\AgdaFunction{velocity}\AgdaSpace{}%
\AgdaKeyword{in}\<%
\\
\>[2]\AgdaOperator{\AgdaFunction{ℚ.-}}\AgdaSpace{}%
\AgdaNumber{1.25}\AgdaSpace{}%
\AgdaOperator{\AgdaDatatype{ℚ.<}}\AgdaSpace{}%
\AgdaSymbol{(}%
\AgdaBound{y}\AgdaSpace{}%
\AgdaOperator{\AgdaFunction{ℚ.+}}\AgdaSpace{}%
\AgdaNumber{2}\AgdaSpace{}%
\AgdaOperator{\AgdaFunction{ℚ.*}}\AgdaSpace{}%
\AgdaFunction{currentSensor}\AgdaSpace{}%
\AgdaBound{x}\AgdaSymbol{)}\AgdaSpace{}%
\AgdaOperator{\AgdaFunction{ℚ.-}}\AgdaSpace{}%
\AgdaFunction{previousSensor}\AgdaSpace{}%
\AgdaBound{x}\AgdaSpace{}\<%
\\
\>[2]\AgdaOperator{\AgdaFunction{×}}\AgdaSpace{}%
\AgdaSymbol{(}%
\AgdaBound{y}\AgdaSpace{}%
\AgdaOperator{\AgdaFunction{ℚ.+}}\AgdaSpace{}%
\AgdaNumber{2}\AgdaSpace{}%
\AgdaOperator{\AgdaFunction{ℚ.*}}\AgdaSpace{}%
\AgdaFunction{currentSensor}\AgdaSpace{}%
\AgdaBound{x}\AgdaSymbol{)}\AgdaSpace{}%
\AgdaOperator{\AgdaFunction{ℚ.-}}\AgdaSpace{}%
\AgdaFunction{previousSensor}\AgdaSpace{}%
\AgdaBound{x}\AgdaSpace{}%
\AgdaOperator{\AgdaDatatype{ℚ.<}}\AgdaSpace{}%
\AgdaNumber{1.25}\<%
\\
\\
\>[0]\AgdaKeyword{abstract}\<%
\\
\>[0][@{}l@{\AgdaIndent{0}}]%
\>[2]\AgdaFunction{safe}\AgdaSpace{}%
\AgdaSymbol{:}\AgdaSpace{}%
\AgdaSymbol{∀}\AgdaSpace{}%
\AgdaBound{x}\AgdaSpace{}%
\AgdaSymbol{→}\AgdaSpace{}%
\AgdaFunction{SafeInput}\AgdaSpace{}%
\AgdaBound{x}\AgdaSpace{}%
\AgdaSymbol{→}\AgdaSpace{}%
\AgdaFunction{SafeOutput}\AgdaSpace{}%
\AgdaBound{x}\<%
\\
\>[2]\AgdaFunction{safe}\AgdaSpace{}%
\AgdaSymbol{=}\AgdaSpace{}%
\AgdaMacro{checkVehicleProperty}\AgdaSpace{}%
\AgdaKeyword{record}\<%
\\
\>[2][@{}l@{\AgdaIndent{0}}]%
\>[4]\AgdaSymbol{\{}\AgdaSpace{}%
\AgdaField{proofCache}%
\>[19]\AgdaSymbol{=}\AgdaSpace{}%
\AgdaString{"path/to/property/file.vclp"}\<%
\\
\>[4]\AgdaSymbol{\}}\<%
\end{code}

%% file: figures/python-training-code.tex
\begin{minted}[autogobble,linenos,python3]{python}
    import vehicle_lang as vcl

    loss_fn = vcl.load_loss_function(
        specification_path="controller-spec.vcl",
        property_name="safe",
        target=vcl.DifferentiableLogic.DL2,
    )

    # Standard code to create and train neural network
    model = ... 
    for epoch in range(num_epochs):
        with tf.GradientTape() as tape:
            loss = loss_fn(controller=model)
        grads = tape.gradient(loss, model.trainable_weights)
        optimizer.apply_gradients(zip(grads, model.trainable_weights))
\end{minted}

%% file: figures/python-output.tex
\begin{minted}[python3]{python}
def lossFn(controller):
  return sample(
    samples=10, 
    domain=[[0.09375,0.09375],[0.90625,0.90625]], 
    body=
      lambda v: 
        sum(0.0, 
          (sum(0.0, -2.75 - controller([xi + 4.0) / 8.0 for xi in v])[0] 
      + 2.0 * v[0] - v[1])) + 
            sum(0.0, - 2.75 + controller([xi + 4.0) / 8.0 for xi in v])[0] 
       + 2.0 * v[0] - v[1]))))
  )
\end{minted}